# Reliability and Performance Assessment of Federated Learning on Clinical Benchmark Data

GeunHyeong Lee[1], Soo-Yong Shin[1,2,*]

## Abstract

As deep learning have been applied in a clinical context, privacy concerns have increased because of the collection and processing of a large amount of personal data. Recently, federated learning (FL) has been suggested to protect personal privacy because it does not centralize data during the training phase. In this study, we assessed the reliability and performance of FL on benchmark datasets including MNIST and MIMIC-III. In addition, we attempted to verify FL on datasets that simulated a realistic clinical data distribution. We implemented FL that uses a client–server architecture and tested client–server FL on modified MNIST and MIMIC-III datasets. FL delivered reliable performance on both imbalanced and extremely skewed distributions (i.e., the difference of the number of patients and the characteristics of patients in each hospital). Therefore, FL can be suitable to protect privacy when applied to medical data.

[1]Department of Digital Health, SAIHST, Sungkyunkwan University, Seoul 06351, Korea, [2]Big Data Research Center, Samsung Medical Center, Seoul 06355, Korea, * Corresponding author, email: sy.shin@skku.edu

# 1. Introduction

Traditional machine learning (ML) and deep learning (DL) require a centralized dataset to train the model. Therefore, they not only require data transfer for collecting data from many devices, persons, or institutions but also have a high computational cost because of training on big datasets. When collecting privacy-sensitive data, such as medical data, privacy protection is a big hurdle. Centralized databases are the main targets of hacking attacks, and the risk of a data breach is severely increased[1, 2]. Moreover, data centralization increases the risk of re-identification of de-identified data with the increased data size[3].

Federated learning (FL) has been suggested to tackle this issue. Originally, FL has been proposed to reduce the computational cost by using the computational cores in mobile devices by Google[4, 5, 6]. In FL, the training is performed on individual clients, and then, the local weights of each client are sent to the server. The server collects the updated local weights and calculates the new global weights. Subsequently, the client downloads the global weights from the server and continues the training process. Many researchers have studied FL in various real-world applications and improved the optimizer and the architecture[7, 8, 9].

This process of FL offers the advantage of personal privacy protection because transferring and centralization of data are not required. This privacy protection concept can be particularly beneficial for medical data analysis (as one of the most sensitive types of personal data). To protect patients' privacy, de-identification methods have been usually applied. However, data centralization is required for both de-identifying data and evaluating the risk of re-identification. If the data are centralized, the risk of a data breach is higher. Moreover, when de-identifying the dataset, we must determine the direct or indirect identifiers in medical data. This is challenging because of a lack of clear guidelines. Health Insurance Portability and Accountability Act (HIPAA) in the U.S. has clear de-identification guidance; it defines 18 types of protected health information to be removed[10]. However, many researchers and social activists claim that the guidance should be revised to enhance privacy protection[11]. On the contrary, FL does not require centralization of raw data. As a result, even the developers of FL cannot access the raw data. Therefore, FL can solve this problem.

In this study, we assess the reliability and performance of FL on two benchmark datasets: the Modified National Institute of Standards and Technology (MNIST) dataset and the Medical Information Mart for Intensive Care-III (MIMIC-III) dataset. We also verify FL in environments that simulated real-world data distributions by modifying MNIST and MIMIC-III datasets.

# 2. Results

## 2.1 MNIST

The proposed approach was evaluated on the MNIST dataset for five different cases (as in Methods). Table 1 presents the values of the area under the receiver operating characteristic curve (AUROC) and F1-score for each case, and Table 2 presents the confusion matrix for each case.

Centralized machine learning (CML) was a baseline training method and set as a control group. CML achieved an AUROC of 0.999 and an F1-score of 0.981. For basic FL, AUROC and F1-score were 0.997 and 0.946, respectively. The initial performance of basic FL was fairly high, with an accuracy of approximately 0.800, which has been continually improved (Supplementary Table 1.A).

Imbalanced FL was designed to reflect a realistic clinical data distribution. As presented in Methods, each client has a different size of training data. Interestingly, the performance of imbalanced FL was significantly superior—AUROC and F1-score of 0.995 and 0.921, respectively. The initial performance was rather poor, as expected. However, after several rounds of processing, the performance has rapidly improved up to an accuracy of 0.900 and slowly increased afterward (Supplementary Table 1.B).

| Experiments | AUROC | F1-score |
|---|---|---|
| CML | 0.999 (0.999, 0.999) | 0.981 (0.978, 0.983) |
| Basic FL | 0.997 (0.996, 0.998) | 0.946 (0.941, 0.950) |
| Imbalanced FL | 0.995 (0.994, 0.995) | 0.921 (0.917, 0.927) |
| Skewed FL | 0.992 (0.991, 0.993) | 0.905 (0.899, 0.911) |
| Imbalanced and skewed FL | 0.990 (0.989, 0.991) | 0.891 (0.884, 0.896) |

**Table 1.** The comparison of experimental results on five different cases of MNIST such as CML (centralized traditional machine learning method), basic federated learning (FL), imbalanced FL, skewed FL, and imbalanced-and-skewed FL. AUROC means the area under the receiver operating characteristic curve. All experiments use the same model and hyperparameters. All results are presented with a 95% confidence interval by resampling the 100 times validation task.

Skewed FL assumed the extreme case. Each client only has one digit from 0 to 9, thereby simulating a situation when each hospital has a unique subpopulation of patients without overlaps. Finally, AUROC and F1-score were 0.992 and 0.905, respectively. As expected, the initial performance was poor; however, it has rapidly improved after the initial rounds. The best performance was achieved in 2,879 rounds (Supplementary Table 1.C).

The most extreme case was designed by combining an imbalanced and a skewed dataset. In this experiment, AUROC and F1-score were 0.990 and 0.891, respectively. Similar to skewed FL, the initial performance was very poor, but it has rapidly improved after the initial rounds (Supplementary Table 1.D).

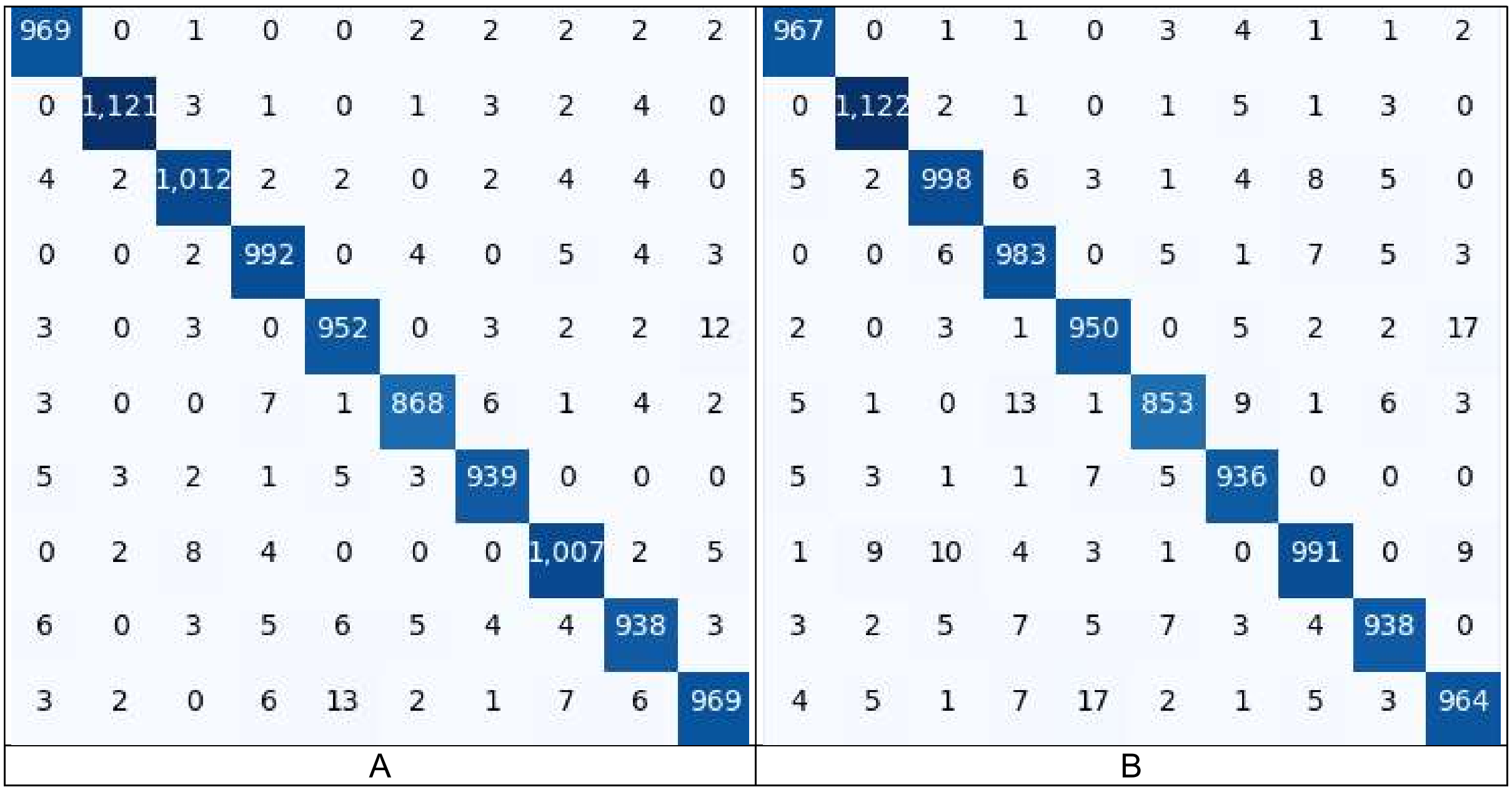

| | | | | | | | | | |
|---|---|---|---|---|---|---|---|---|---|
| 960 | 0 | 1 | 1 | 0 | 7 | 8 | 1 | 2 | 0 |
| 0 | 1,112 | 5 | 3 | 0 | 2 | 3 | 1 | 9 | 0 |
| 10 | 3 | 935 | 26 | 11 | 3 | 8 | 13 | 20 | 3 |
| 5 | 1 | 15 | 917 | 1 | 28 | 1 | 12 | 22 | 8 |
| 1 | 5 | 7 | 1 | 903 | 0 | 15 | 3 | 4 | 43 |
| 9 | 4 | 6 | 33 | 10 | 778 | 12 | 6 | 18 | 16 |
| 24 | 1 | 14 | 0 | 13 | 14 | 889 | 0 | 3 | 0 |
| 1 | 17 | 25 | 6 | 7 | 0 | 0 | 946 | 2 | 24 |
| 6 | 8 | 15 | 21 | 16 | 16 | 12 | 6 | 867 | 7 |
| 5 | 9 | 1 | 8 | 34 | 13 | 1 | 26 | 6 | 906 |

C

| | | | | | | | | | |
|---|---|---|---|---|---|---|---|---|---|
| 967 | 0 | 0 | 1 | 0 | 2 | 8 | 1 | 1 | 0 |
| 0 | 1,107 | 4 | 2 | 0 | 1 | 4 | 2 | 15 | 0 |
| 14 | 12 | 901 | 10 | 9 | 2 | 19 | 18 | 42 | 5 |
| 6 | 2 | 30 | 875 | 0 | 33 | 4 | 24 | 26 | 10 |
| 4 | 1 | 12 | 0 | 888 | 0 | 11 | 3 | 12 | 51 |
| 17 | 8 | 8 | 41 | 11 | 724 | 24 | 13 | 37 | 9 |
| 11 | 3 | 6 | 0 | 7 | 6 | 921 | 1 | 3 | 0 |
| 3 | 10 | 31 | 2 | 3 | 0 | 0 | 949 | 3 | 27 |
| 12 | 12 | 13 | 17 | 9 | 26 | 18 | 15 | 841 | 10 |
| 10 | 8 | 2 | 7 | 39 | 6 | 1 | 47 | 10 | 879 |

D

| | | | | | | | | | |
|---|---|---|---|---|---|---|---|---|---|
| 957 | 0 | 2 | 3 | 0 | 3 | 14 | 1 | 0 | 0 |
| 0 | 1,078 | 7 | 4 | 0 | 2 | 4 | 1 | 39 | 0 |
| 19 | 10 | 899 | 8 | 7 | 2 | 34 | 9 | 33 | 11 |
| 8 | 2 | 29 | 883 | 1 | 23 | 4 | 18 | 35 | 7 |
| 5 | 0 | 2 | 0 | 863 | 1 | 39 | 5 | 8 | 59 |
| 30 | 9 | 2 | 65 | 7 | 682 | 30 | 17 | 41 | 9 |
| 10 | 3 | 4 | 1 | 9 | 9 | 917 | 1 | 4 | 0 |
| 2 | 18 | 30 | 5 | 7 | 0 | 0 | 931 | 4 | 31 |
| 20 | 6 | 7 | 24 | 9 | 28 | 22 | 10 | 835 | 13 |
| 15 | 8 | 0 | 10 | 49 | 7 | 2 | 46 | 12 | 860 |

E

**Table 2.** Confusion matrices for MNIST experiments. (A) Centralized machine learning (CML), (B) Basic FL. (C) Imbalanced FL. (D) Skewed FL. (E) Imbalanced and Skewed FL.

### 2.2 MIMIC-III

The proposed approach was evaluated on MIMIC-III in two different cases to compare the performance with a reported benchmark. FL experiments were performed on three individual clients. Apart from AUROC and F1-score, we also refer to the area under the precision–recall curve (AUPRC), which is reported in the benchmark[12]. The results are presented in Tables 3–4 and Supplementary Figure 1.

The basic MIMIC-III FL experiment has been performed as follows. The original dataset has been split into three datasets without duplication, and each client has been trained on only one of these datasets. In this MIMIC-III experiment, AUROC and F1-score were 0.850 and 0.944, respectively. This is comparable with the results obtained by Harutyunyan et al[12], who reported an AUROC of 0.855.

| Experiments | AUROC | F1-score | AUPRC |
|---|---|---|---|
| Benchmark | 0.855 (0.835, 0.873) | - | 0.485 (0.431, 0.537) |
| Basic FL | 0.850 (0.830, 0.869) | 0.944 (0.938, 0.950) | 0.483 (0.427, 0.537) |
| Imbalanced FL | 0.850 (0.829, 0.869) | 0.943 (0.937, 0.949) | 0.481 (0.426, 0.535) |

**Table 3.** Results for the benchmark, basic federated learning (Basic FL), and Imbalanced FL on MIMIC-III. All results are presented with a 95% confidence interval by resampling the 10,000 times validation task.

The imbalanced FL experiment is an extension of the basic MIMIC-III FL. All data were split without duplication into 50%, 30%, and 20% to be assigned to separate clients. The rounds progressed up to 30. As a result, AUROC and F1-score were 0.850 and 0.943, respectively.

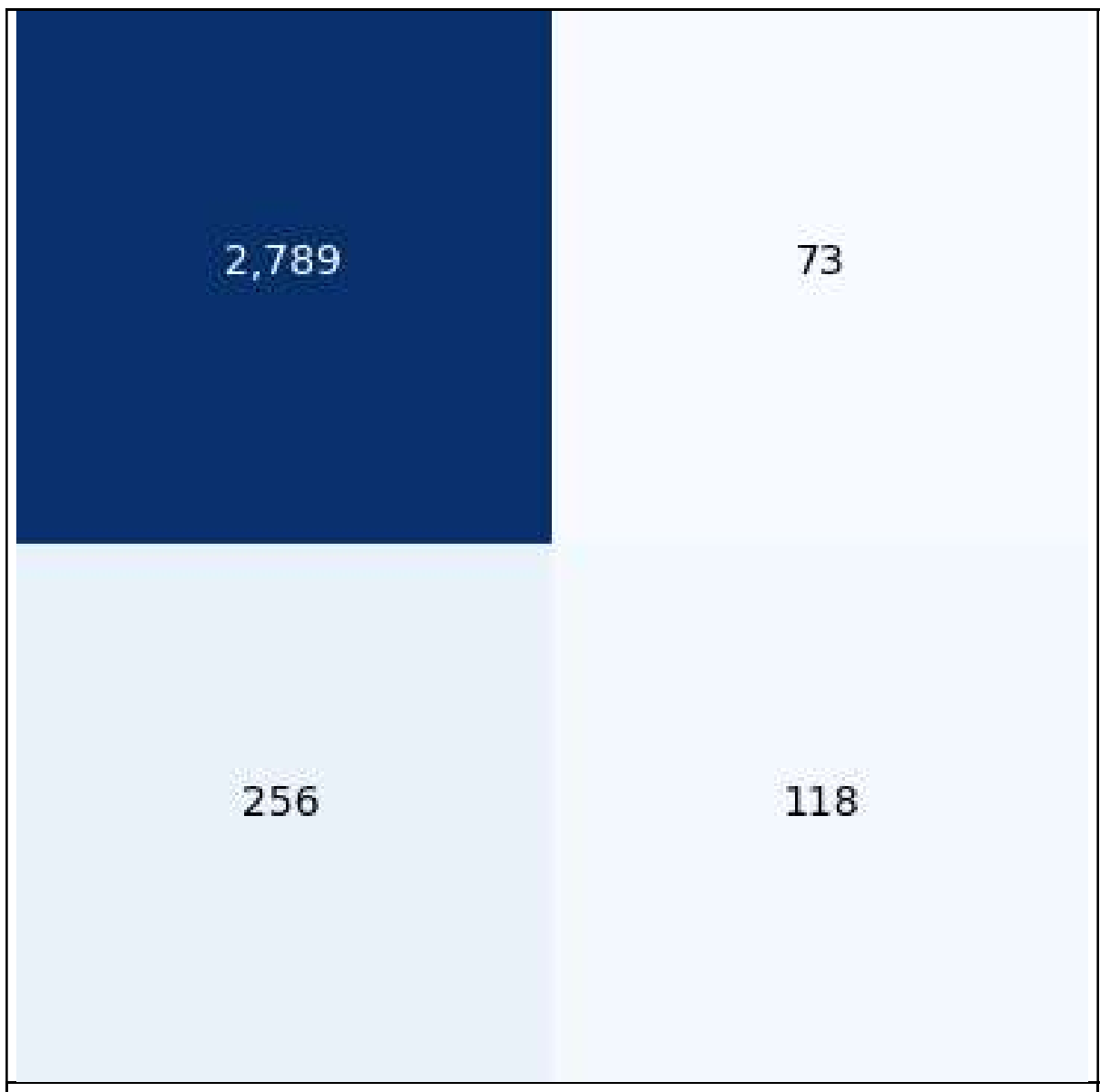

Basic FL

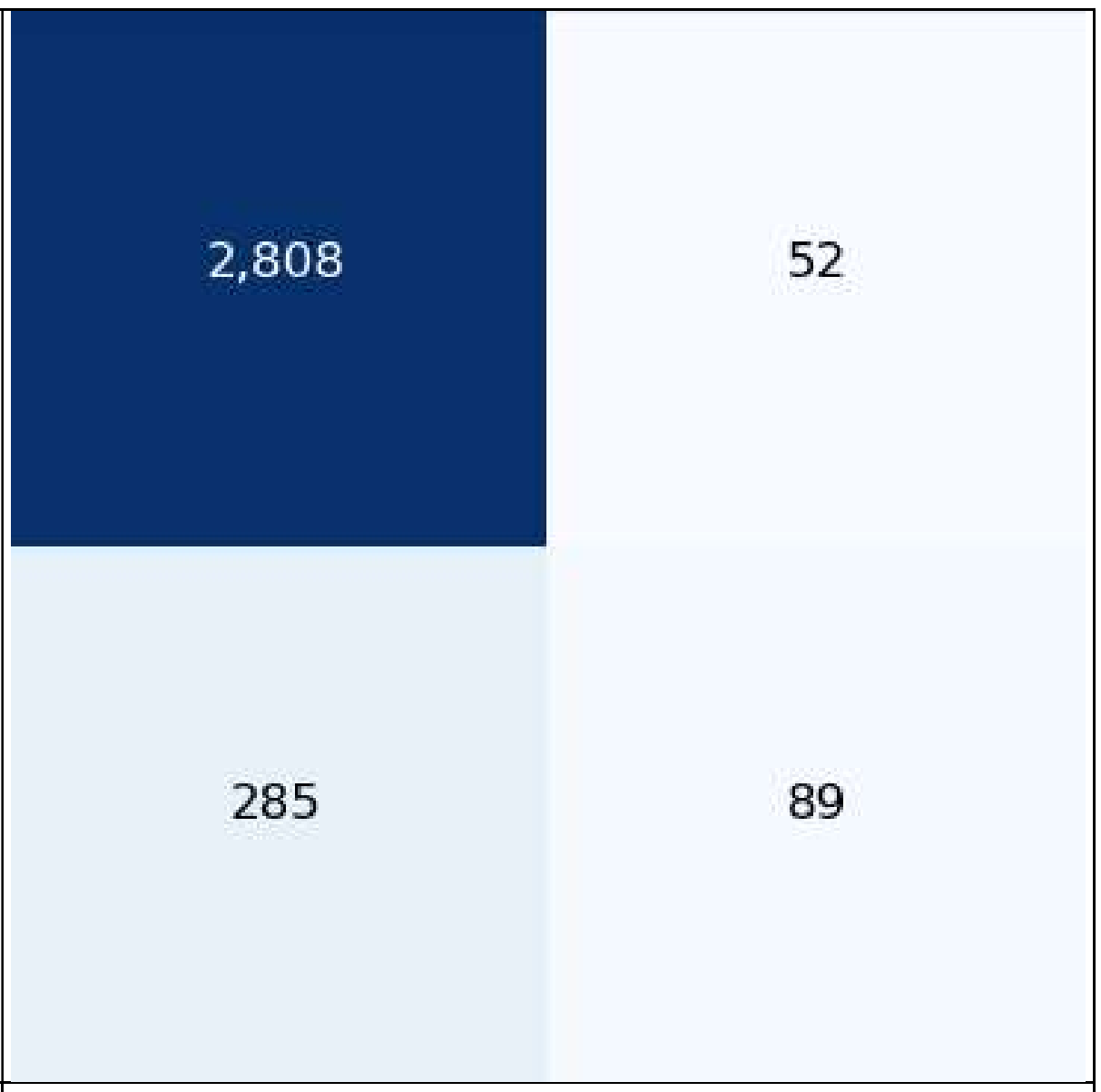

Imbalanced FL

**Table 4.** Confusion matrix of federated learning (FL) on MIMIC-III. (A) Basic FL. (B) Imbalanced FL.

## 3. Discussion

When comparing the performances of CML and FL in basic MNIST experiments, both AUROC and F1-score are high. Unexpectedly, when using an imbalanced dataset, FL delivered good performance with only small differences (AUROC and F1-score of 0.003 and 0.035, respectively). When using a skewed dataset, FL also yielded remarkable results for AUROC and F1-score. When comparing the confusion matrices for experiments with four datasets (i.e., normal, imbalanced, skewed, and a combination of two distributions), FL showed some deterioration for similarly looking numbers (e.g., 3 vs. 5; 4 vs. 9). Even in the basic MNIST classification, the performance was relatively poor in these cases. However, this problem was not related to small sizes of training datasets. When we monitored the size of training datasets of each client, the dataset for class 5 was not small. Moreover, depending on the experiment, the datasets for classes 1 or 7 could be small, but superior classification performance has been achieved. This trend has been maintained in the experiments with basic FL and imbalanced FL using the MIMIC-III dataset.

The FL experiments using MIMIC-III also exhibited good and competitive performance, compared to a benchmark that has been trained on CML. The experimental results of in-hospital mortality using MIMIC-III dataset, which is a well-known dataset with real clinical data, has also showed good performance. This experiment was performed by splitting the randomly selected MIMIC-III data into three parts (i.e., from the perspective of each institution, learning on one-third of total data). However, performances of FL and CML were almost the same, with only a difference of 0.005 in AUROC, compared to state-of-the-art (SOTA) reported by Harutyunyan et al[12]. Before the experiments, we expected that the performance of FL would be slightly inferior to that of CML because FL uses a distributed dataset instead of a centralized dataset. Nevertheless, no significant difference was found in well-known evaluation indicators such as accuracy, sensitivity, precision, and F1-score (except for AUROC). Experimental results with an imbalanced dataset were very similar as those for basic FL. Therefore, an individual client may only use a small amount of data for training in FL, but the resulting effect is similar to training on all data.

The performance of FL was verified using two datasets with changed data distribution: imbalanced (with unproportionally represented classes) and skewed (the distribution of the target variable is different), to imitate real-world medical data. As a result, FL was comparable to CML. During the initial rounds, only a relatively small amount of data was used on each client instead of an ensemble, so that the performance of FL was significantly inferior to that of CML. However, in the subsequent rounds, the performance of FL (with respect to AUROC and F1-score) becomes close to that of CML. Typically, medical centers have datasets with very different distributions, and our results demonstrate that FL is suitable for real-world medical datasets without data centralization.

The reason why FL delivered comparable performance might be that the weight updates, and the process of federated averaging (FedAVG)[4] could have a similar effect in mini-batches[13, 14, 15] and ensembles[16]. In FL, each client trains on relatively small data and then transfers the local weights to the server. Next, the server collects the local weights and updates the global weights that reflect all the data through FedAVG. Subsequently, the round is repeated to improve the global weights. Hence, individual clients are an element of a mini-batch, and FedAVG is similar to ensemble processing. When implementing FL, we used the widely known FedAVG aggregation method[5], but it does not guarantee the best choice. To solve this problem, many researchers have studied aggregation methods that can work well with abnormal distributions such as FedProx[17], FSVRG[18], CO-OP[19], LoAdaBoost FedAVG[20], and RFA[21]. Hyperparameter selection requires further research.

As a summary, our experiments have demonstrated the potential of FL in terms of performance and data protection, which is important for dealing with sensitive medical data. Specifically, in FL, only weights are transferred, and the participants are unaware of each other's local datasets. This can prevent leaks of personal information. In addition, the proposed approach can be used to supplement existing ones and to avoid problems that may occur during the de-identification process. The future direction of research is to use FL for actual medical data through collaborations with multiple institutions. Tasks such as expanding the client–server version of FL and improving communication will be expected to be important for using FL in real-world medical data with multiple institutions.

## 4. Methods

Currently, FL is supported by several open-source projects, including TensorFlow Federated (TFF) in TensorFlow 2.0[22], PySyft[23, 24], and Federated AI Technology Enabler (FATE)[25, 26]. However, all these libraries have some limitations on stability or extensibility. We implemented our own client-server version of FL using Python to directly control the learning/communication process and to reflect the characteristics of the dataset.

### 4.1 Server

The server was developed using Django framework and Python in Amazon Web Services (AWS). The server provides several application programming interfaces (APIs) for communication with the client, as shown in Table 4, and performs FedAVG, which calculates the weight averages. FedAVG is a widely used optimization algorithm that calculates the average value when the local weights collected from the client reach a specific amount. The implemented code was deployed and managed in AWS Beanstalk, which has been continuously monitored during the process of training.

| Method | URL | Parameter | Description | Return |
|---|---|---|---|---|
| GET | /round | | Request the current round | Number |
| GET | /weight | | Request the global weights | List |
| PUT | /weight | List | Update the local weights | |

**Table 5.** API calls provided by the server

### 4.2 Client

The client consists of three components. The first one is the local learning component of each client, which builds a suitable model for the dataset in the learning phase. The second one is the communication component, which updates local weights according to the results of local training (the first component) to the server and downloads the global weights from the server. The third one is the performance measure component. Using the downloaded global weights, the performance of each client was measured. The implemented code was deployed on an AWS EC2 instance. We used specifications of g4dn.xlarge, which uses NVIDIA T4 Tensor core GPU, on the Amazon instance.

### 4.3 Communications

The client–server communication for FL was implemented based on McMahan et al.'s ideas[27]. However, the implemented code has some differences. The communication assumes that all clients (hospitals) are always powered (as a typical computer, not a mobile device) and the online status is maintained by a wired network connection. In addition, rather than selecting clients by the eligibility criteria from multiple client pools (thousands or millions), the code was implemented to manage a predefined fixed number of clients. In other words, all clients can participate in the round.

The schematic diagram of client–server communication of FL is shown in Supplementary Figure 2. The client decides whether to participate in the current round through the API. If it has already participated (sending local weights to the server), it waits to participate in the next round. The server waits for the client's weight updates and ensures that no clients are eventually dropped out. All communications are performed through the API provided by the server. The monitoring system was used to continuously observe system abnormalities.

### 4.4 MNIST

The MNIST dataset contained 70,000 samples (including 60,000 for training and 10,000 for testing). The basic model was a simple artificial neural network (ANN) with an input layer, one hidden layer with 128 units with the rectified linear unit (ReLU) activation function, and an output layer. The hyperparameters for training were set as follows: batch size 32, maximum 1,000 epochs, and early stopping. Stochastic gradient descent (SGD) was used as an optimizer[28].

For FL, we used 10 individual clients as in a real environment. We modified the datasets and hyperparameters of the learning algorithms. The datasets were modified considering the difference in distribution of medical data between hospitals. Hyperparameters were adjusted for training in each client. The proposed approach was evaluated on the MNIST dataset in four different experiments.

1. **Basic federated learning**: this experiment is to check the basic performance of the FL. Ten clients have randomly selected 600 images from the basic dataset. We continued the progress up to 500 rounds and observed the results.
2. **Imbalanced federated learning**: this experiment is an extension of basic MNIST FL. Other

environments are the same, but only one is different. Each client uses different sizes of randomly selected data for training.

3. **Modified federated learning with a skewed distribution**: the MNIST dataset was split into single-digit groups, from 0 to 9. Each of the 10 numbers was assigned to 10 different clients. Consequently, each client had a single digit instead of 10. This modified MNIST simulates an extremely skewed distribution. Each client randomly selects 600 images from a dataset that has a single digit for training. The same simple ANN was used in the experiments as in the basic model. The hyperparameters were set as follows: five epochs and the batch size of ten. We continued the progress up to 3,000 rounds and observed the results. For the evaluation, a model was created with the latest updated global weights, and 10,000 test samples were used.

4. **Modified federated learning for imbalanced and skewed distributions**: this experiment is an extension of the modified MNIST FL that represents skewed distribution. Each client was trained on data with imbalanced and skewed distribution. Hence, each client was trained only on a single digit using a randomly selected sample.

### 4.5 MIMIC-III

We conducted the experiments on MIMIC-III with three individual clients. As a reference, we used a SOTA benchmark[12]. The MIMIC-III dataset was used to predict in-hospital mortality. After the preprocessing, the dataset contained 21,139 samples (including 17,903 for training and 3,236 for testing). The basic model was a standard long short-term memory (LSTM) with reference to the benchmark[12]. The LSTM was chosen with 16 hidden units, depth 2, dropout 0.3, timestep 1.0, batch size 8, and an adaptive moment estimation (ADAM) optimizer. We compared the results of our experiments with the results by Harutyunyan et al.[12], who performed the benchmark experiment. All experiments were evaluated using AUROC and F1-score.

For FL, randomly chosen samples from the original dataset were divided into three datasets without duplication and assigned to each client. This simulates having three different institutions. The same basic LSTM was used, and hyperparameters were set as follows: two epochs and the batch size of four. We continued the progress up to 30 rounds and observed the results. All experiments were evaluated using F1-score and AUROC. The proposed approach was evaluated on the MIMIC-III dataset in two different experiments.

1. **Basic federated learning**: The basic MIMIC-III FL experiment was performed so that each client has been trained on a subset of data that were split into three parts without duplication.
2. **Imbalanced federated learning**: This experiment is an extension of basic MIMIC-III FL. However, all data was split without duplication into 50%, 30%, and 20%, respectively, and was assigned to each client.

### 4.6 Evaluation

All experiments compared the performance of training on imbalanced and/or skewed distribution on each client versus normal distribution. In addition, all experiments evaluated AUROC, F1-score, and the confusion matrix. Finally, we used bootstrapping to compute significant differences between the experiments. We calculated 95% confidence interval and resampled the test set K times (MNIST K is 100, while MIMIC-III K is 10,000), using 2.5 and 97.5 percentiles of scores.

## 5. Data Availability

MNIST dataset analyzed during the current study is available in the Keras package in the TensorFlow framework. MIMIC-III dataset analyzed during the current study is available on the PhysioNet repository[29]. The preprocessed dataset used in this study is available on mimic3-benchmarks repository[30].

## 6. Code Availability

The implemented server code is available on FL_Server repository at https://github.com/bmiskkuedu/FL_Server and the client code is available on FL_Client repository at https://github.com/bmiskkuedu/FL_Client. Additionally, the original code to generate the MIMIC-III experiment used in this study refer to mimic3-benchmarks repository[30].

## 8. Acknowledgements

This work was supported by the Institute for Information & Communications Technology Promotion (IITP) grant funded by the Korea government (MSIT) (2018-0-00861, Intelligent SW Technology Development for Medical Data Analysis).

## 9. Author contributions

G.H. implemented the code and the framework. S.Y designed the experiments and supervised the study. All authors wrote the manuscript and discussed the results.

## 10. Competing interests

The authors declare no competing interests.

## 11. Additional information

**Supplementary information**

A

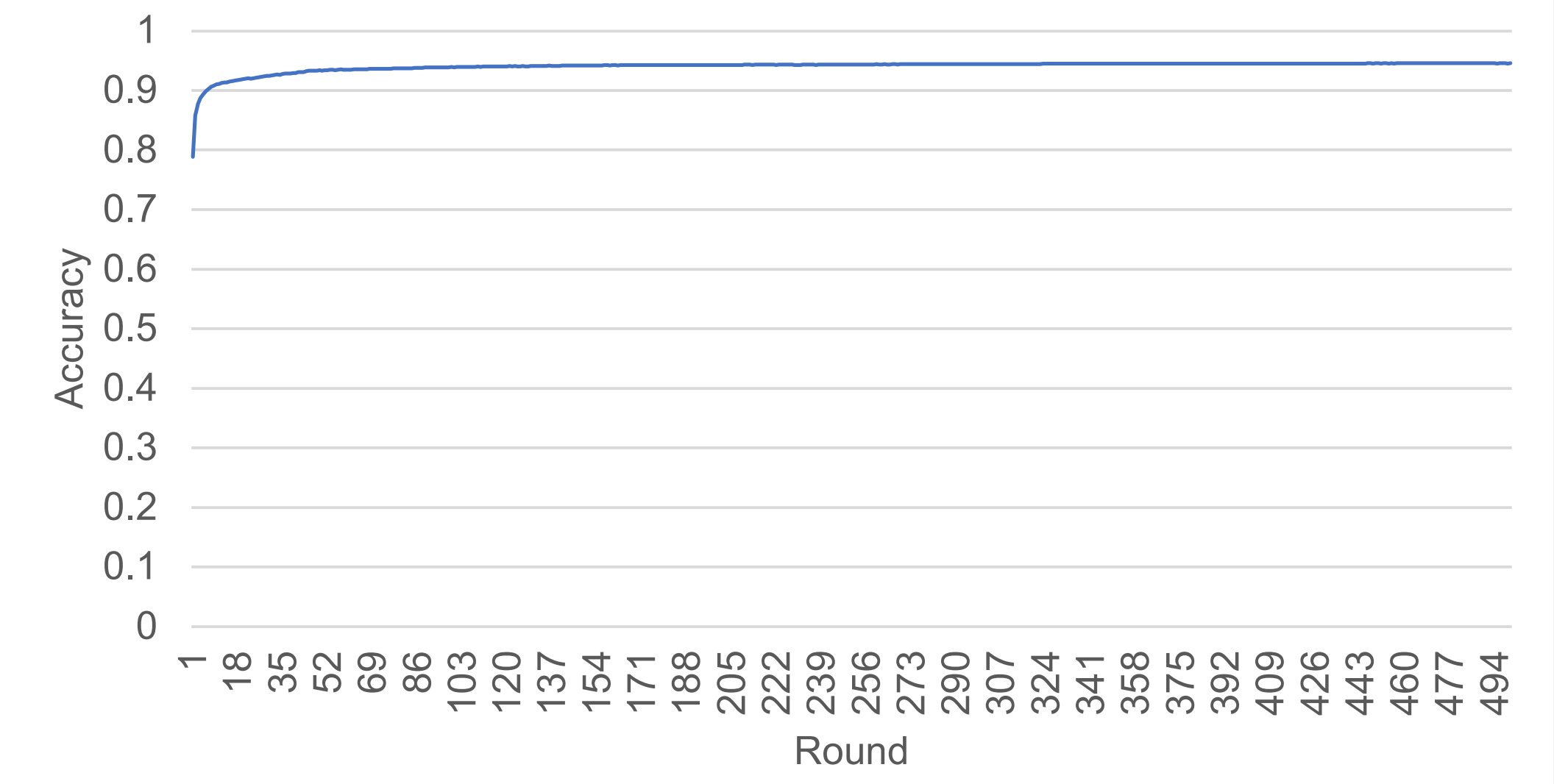

B

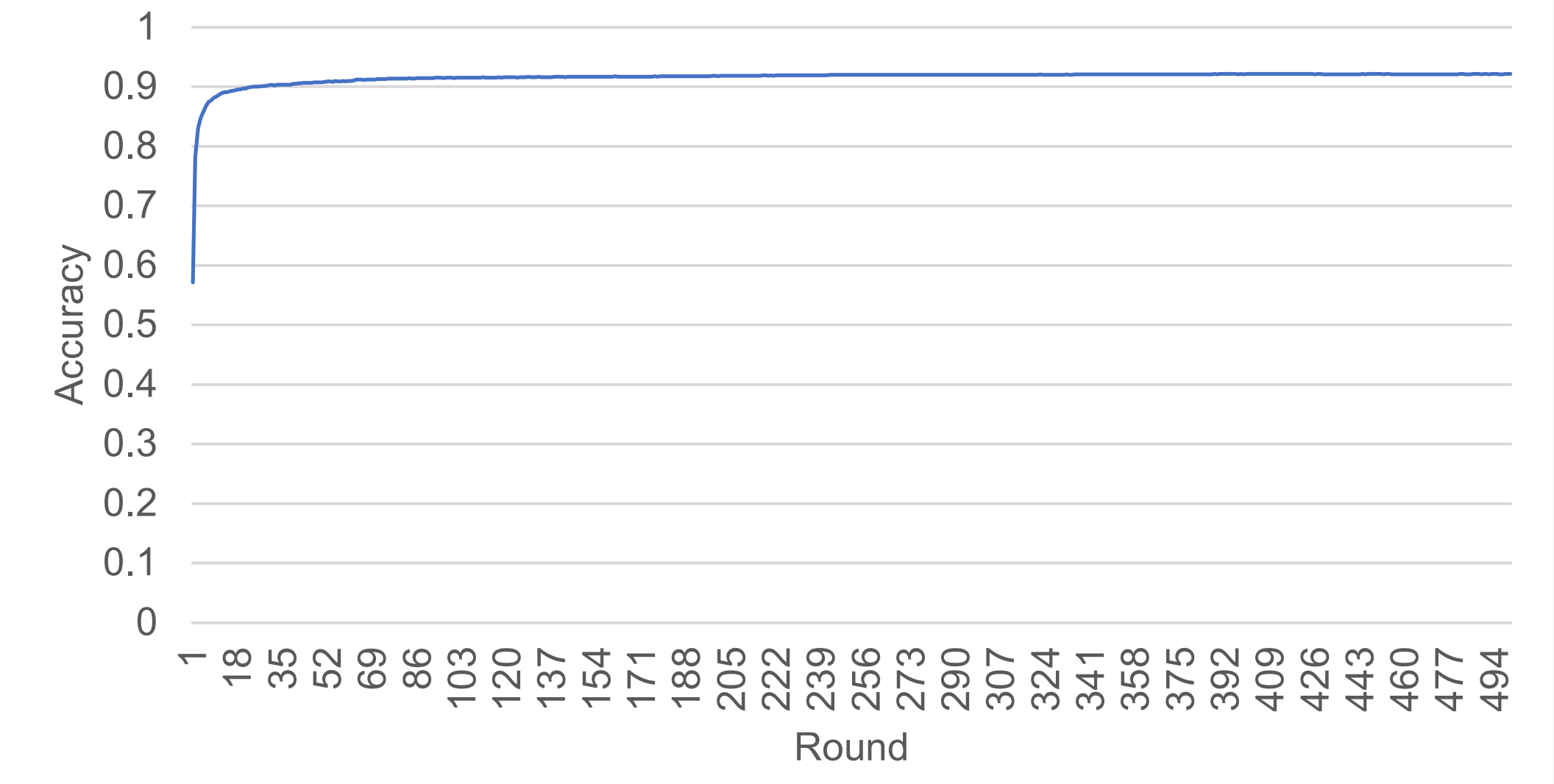

C

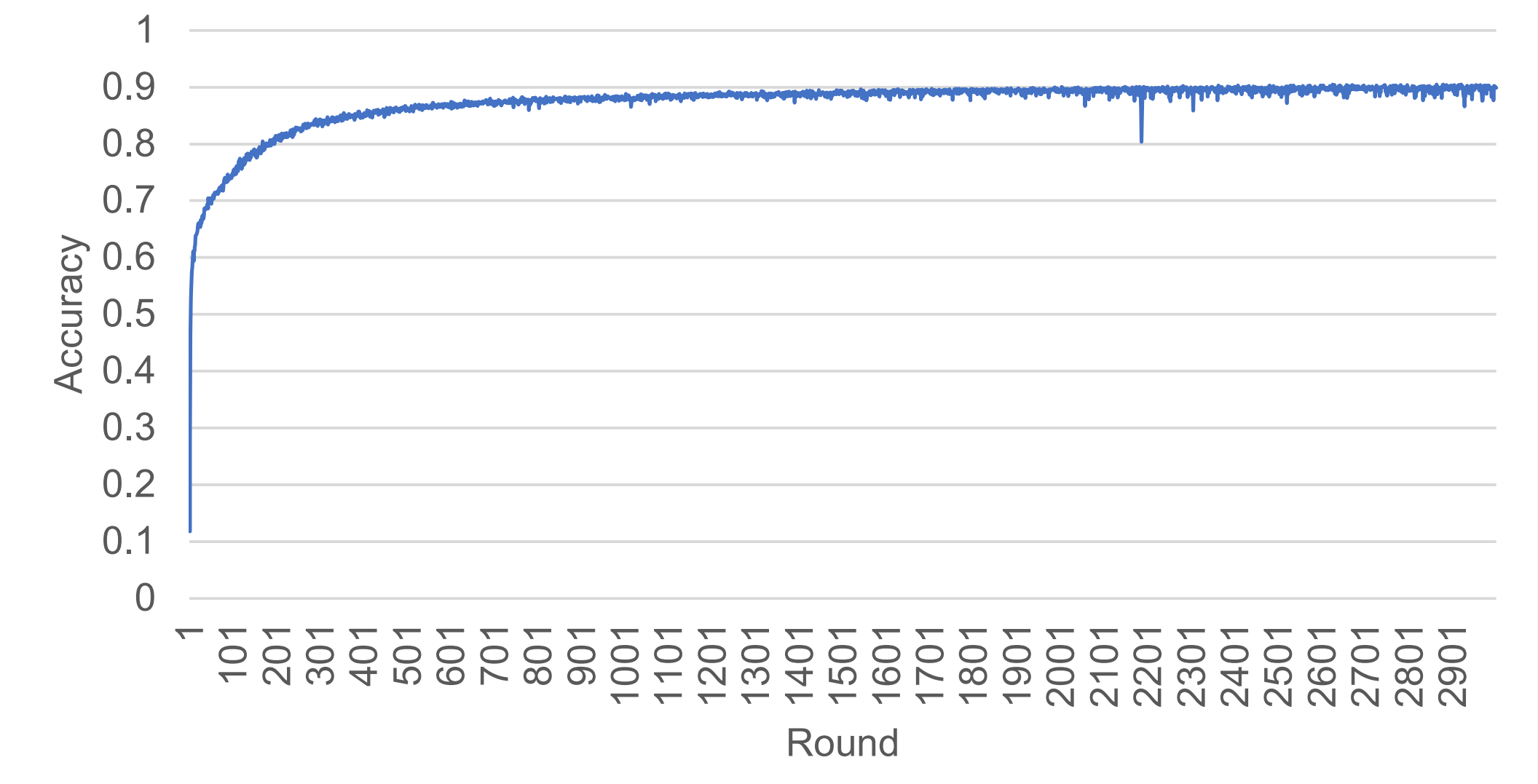

| | |
|---|---|
| D | 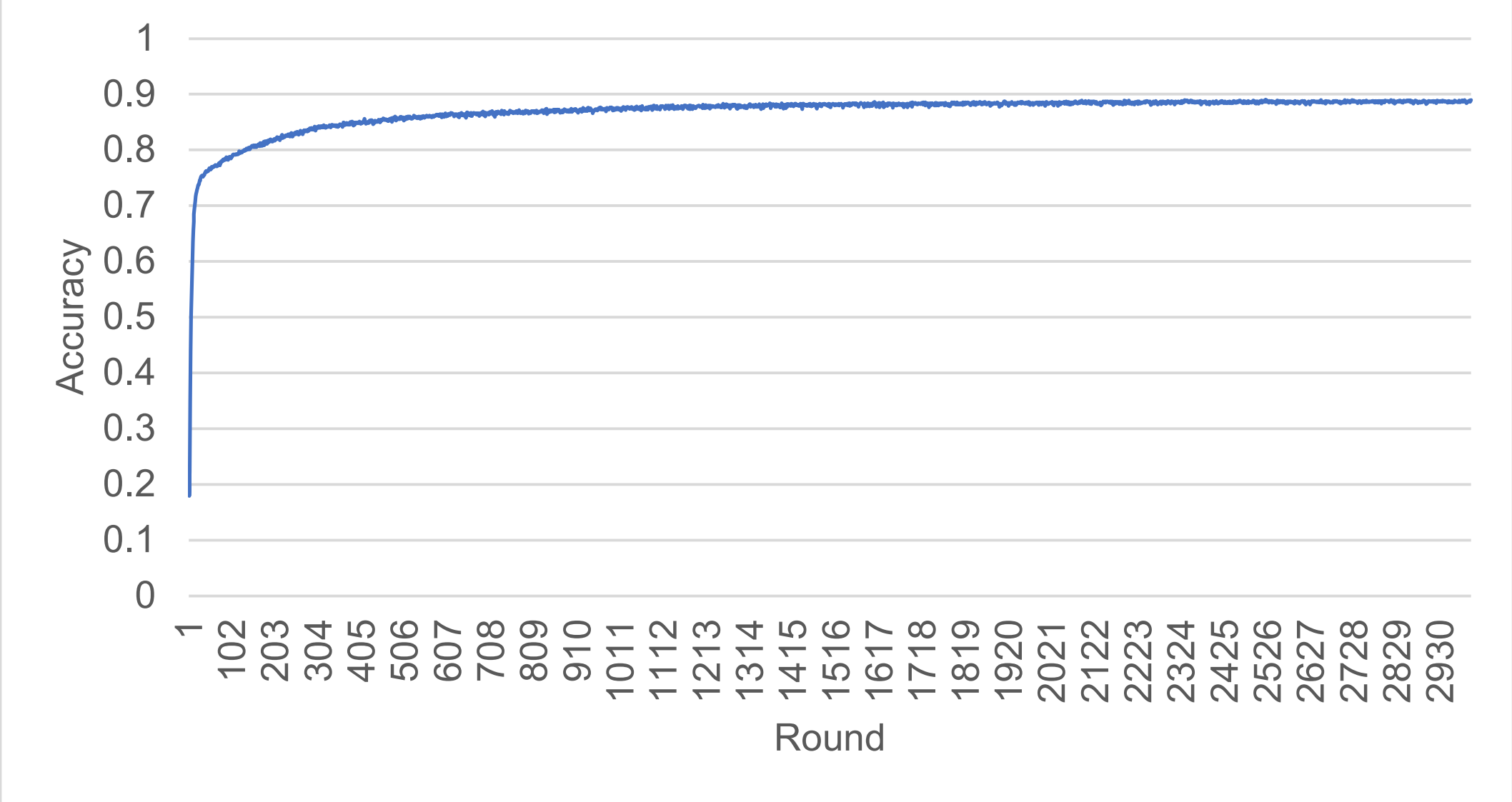 |

**Supplementary Table 1.** Accuracy changes on each round of MNIST federated learning (FL) experiments. (A). Basic FL. (B) Imbalanced FL. (C) Skewed FL. (D) Imbalanced and skewed FL.

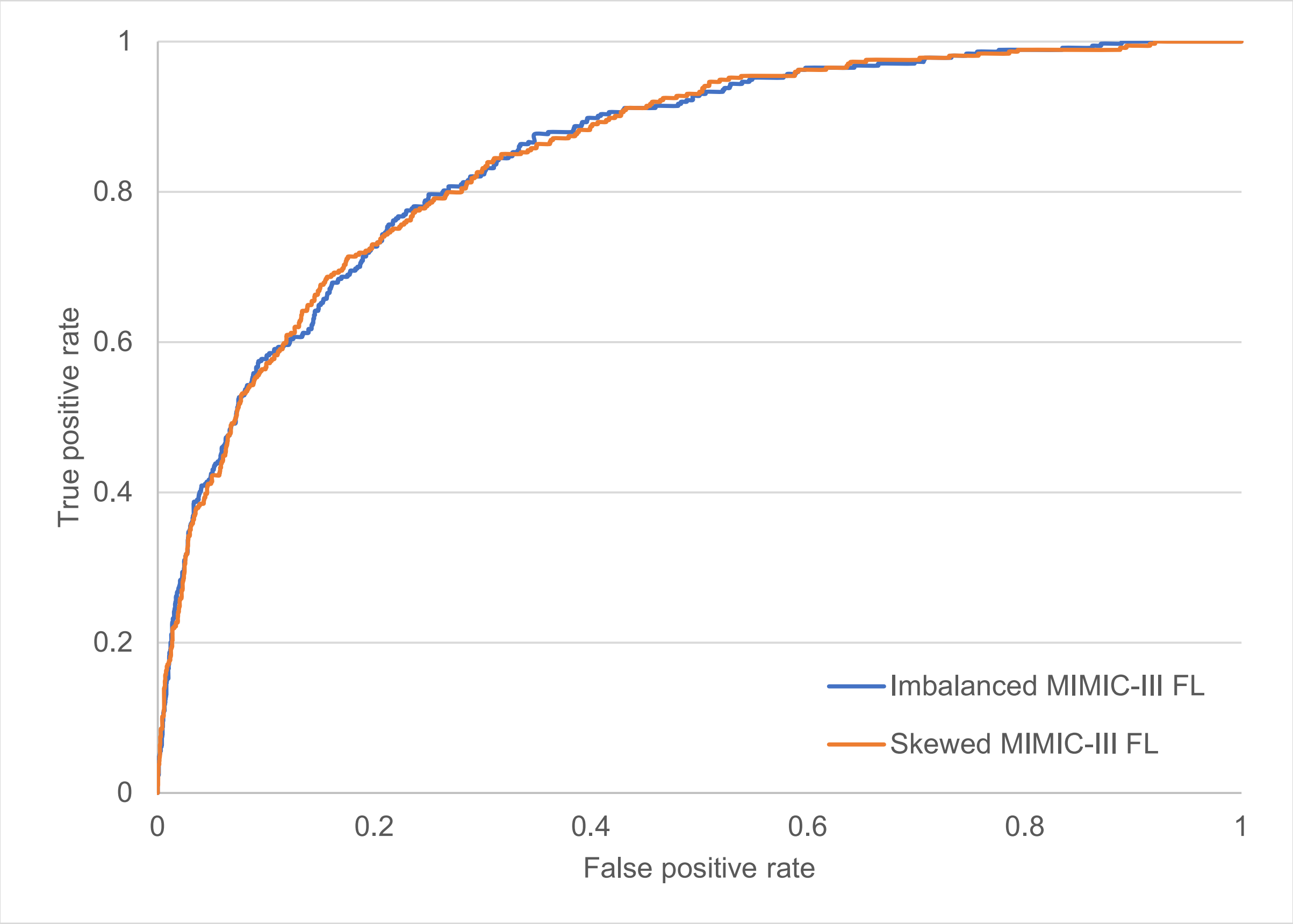

**Supplementary Figure 1.** Area under the receiver operating characteristic curve for each experiment using the MIMIC-III dataset.

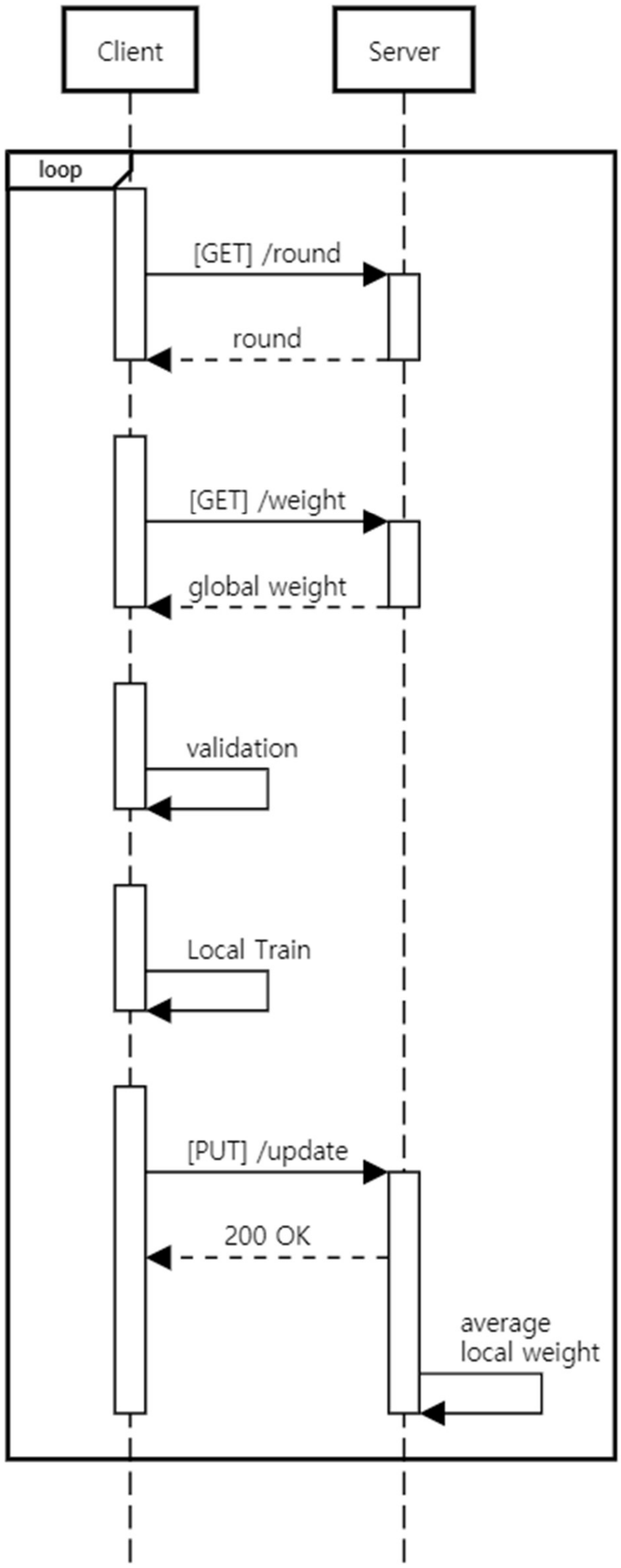

**Supplementary Figure 2.** Client-server version of the communication logic in federated learning.